%% file: main.tex
\pdfoutput=1

\documentclass[11pt]{article}

\usepackage{ACL2023}

\usepackage{times}
\usepackage{latexsym}

\usepackage[T1]{fontenc}

\usepackage[htt]{hyphenat}

\usepackage[utf8]{inputenc}

\usepackage{microtype}

\usepackage{inconsolata}

\usepackage{graphicx}

\title{Enhancing Pipeline-Based Conversational Agents with Large Language Models }

\author{Mina Foosherian \\
  BIBA - Bremer Institut für \\ Produktion  und Logistik GmbH \\
  Hochschulring 20 \\
  28359, Bremen, Germany \\
  \texttt{fos@biba.uni-bremen.de} \\\And
  Hendrik Purwins \\
  Accenture GmbH\\
  Center of Excellence for LLMs\\ 
  Balanstrasse 73 \\
  81541, Munich, Germany \\
  \texttt{hendrik.purwins@accenture.com} \\\AND
  Purna Rathnayake \\
  University of Bremen \\
  Faculty of Physics and Electrical Engineering \\
  Otto-Hahn-Allee 1\\
  28359, Bremen, Germany \\ 
  \texttt{purnasam@uni-bremen.de} \\\And
  Touhidul Alam \\
  Accenture GmbH\\
  Liquid Studio \\
  Balanstrasse 73 \\
  81541, Munich, Germany \\
  \texttt{touhidul.alam@accenture.com} \\\AND
    Rui Teimao \\
    Lloyds Banking Group\\
    Conversational Banking Lab\\
    125 London Wall, Barbican \\
    London EC2Y 5BL, England \\
\texttt{rui.teimao@lloydsbanking.com} \\\And
  Klaus-Dieter Thoben \\
  University of Bremen \\
  Faculty of Production Engineering \\
  Badgasteiner Str. 1\\
  28359, Bremen, Germany \\
  \texttt{tho@biba.uni-bremen.de} \\
  }

\begin{document}
\maketitle
  \footnotetext{The fifth author R.T. was at Accenture during the time of the research. This paper has been accepted at the TamingLLMs Workshop at SigDial 23 in Prague, Sept. 12th 2023.}

\begin{abstract}\footnote{}
The latest advancements in AI and deep learning have led to a breakthrough in large language model (LLM)-based agents such as GPT-4. However, many commercial conversational agent development tools are pipeline-based and have limitations in holding a human-like conversation. 
This paper investigates the capabilities of LLMs to enhance pipeline-based conversational agents during two phases: 1) in the design and development phase and 2) during operations. In 1) LLMs can aid in generating training data, extracting entities and synonyms, localization, and persona design. In 2) LLMs can assist in contextualization, intent classification to prevent conversational breakdown and handle out-of-scope questions, auto-correcting utterances, rephrasing responses, formulating disambiguation questions, summarization, and enabling closed question-answering capabilities. We conducted informal experiments with GPT-4 in the private banking domain to demonstrate the scenarios above with a practical example. 
Companies may be hesitant to replace their pipeline-based agents with LLMs entirely due to privacy concerns and the need for deep integration within their existing ecosystems.
A hybrid approach in which LLMs' are integrated into the pipeline-based agents allows them to save time and costs of building and running agents by capitalizing on the capabilities of LLMs while retaining the integration and privacy safeguards of their existing systems. 
\end{abstract}

\input{sections/01-introduction}

\input{sections/02-state_of_the_art}
\input{sections/03-approach}
\input{sections/04-experiments}
\input{sections/05-outlook}

\section*{Acknowledgements}
This work was supported by the European Union’s Horizon 2020 research and innovation program via the project COALA “COgnitive Assisted agile manufacturing for a LAbor force supported by trustworthy Artificial Intelligence” (Grant agreement 957296). In addition, this work was supported by REasoning for Conversation and Information Technology Exchange (RECITE) project which is an OASIS Open Project dedicated to developing a standard for dialog modeling in conversational agents.

\section*{Disclaimer} \label{sec:disclaimer}
This document is intended for general informational purposes only and does not take into account the reader’s specific circumstances, and may not reflect the most current developments. In particular, this research paper does not take into account the specific needs of an IT ecosystem and network, which may vary and require unique action. The reader should independently assess their specific needs in deciding to use any of the tools mentioned.  Google Dialogflow, Cognigy, Rasa, etc. tools are not Accenture tools.  Accenture does not make any representation that it has vetted or otherwise endorses these tools.  Accenture disclaims any liability for their use, effectiveness or any disruption or loss arising from the use of these tool. Accenture disclaims, to the fullest extent permitted by applicable law, any and all liability for the accuracy and completeness of the information in this paper and for any acts or omissions made based on such information. Accenture does not provide legal, regulatory, audit, or tax advice. Readers are responsible for obtaining such advice from their own legal counsel or other licensed professionals.  
\bibliography{custom}
\bibliographystyle{acl_natbib}
\appendix


\input{sections/06-appendix}

\end{document}

%% file: sections/01-introduction.tex
\section{Introduction}
The field of conversational artificial intelligence (CAI) has experienced significant advances in recent years, with the emergence of both commercial and open-source CAI development platforms such as Google Dialogflow, Amazon's Alexa Skills Kit, Cognigy, and Rasa, as well as the more recent large language model (LLM)-based conversational agents (CA) like ChatGPT.

CAs can be text-based agents (Chatbots), Voice-User interfaces (VUI), or embodied-dialog Agents (EDA)~\citep{Harms.2019} and generally aim to replace or empower humans through natural language interaction.

CAs can be pipeline-based or end-to-end~\citep{Chen_2017}. In pipeline-based CAs, the natural language understanding (NLU) component processes the user's message sequentially to identify their goal (intent recognition), and extract information pieces called entities. The dialog management component tracks the dialog state and decides on the next action based on the current state. Finally, the natural language generation (NLG) component builds and returns the response. The CAs ``intelligence'' relies on the agent's training data and internal logic used to create its NLU and dialog management models~\citep{Harms.2019}.

The end-to-end CAs rely on dialog models trained with large training datasets~\citep{Chen_2017}. These models learned hidden relations between input and output utterances, effectively avoiding that developers create interim representations~\citep{dinan2021anticipating}. A downside is that the necessity of larger datasets makes end-to-end approaches less applicable in domains such as manufacturing, where developers cannot derive training data from existing human-human conversations. End-to-end CAs also bear substantial safety issues, such as generating offensive language and responding inappropriately to offensive content or in safety-critical situations~\citep{dinan2021anticipating}. Combinations of pipeline-based and end-to-end approaches are also feasible. Rasa Open Source, for instance, already supports both~\citep{endtoendrasa2, intentless-rasablog}.

This article uses the term LLM to refer to language models trained with an end-to-end approach on a large amount of high-quality training data. Prominent LLMs comprise GPT-4 by OpenAI, PaLM by Google, and LLaMA by Meta AI. Such models can, for instance, possess emergent abilities, be hard to steer, and humans will likely have difficulties interpreting how they work~\citep{bowman2023things}.


In this article, we demonstrate how LLMs can expand the capabilities of pipeline-based CAs without removing the pipeline altogether. The impact of LLMs helps the pipeline-based CAs in generating training data for intent classification, the identification of domain-specific entities and synonyms, requirement characterization for the agent and its personalization and localization, among others. During deployment, LLMs can provide auto-correction to user input, handle context switching and out-of-scope questions, introduce response variability, create conversation summarizations and perform closed Question-Answering (Q\&A). \footnote{Please refer to Section~\ref{sec:disclaimer} for a disclaimer. }



%% file: sections/02-state_of_the_art.tex
\section{The State of Conversational Agents}

Broadly, CAs can be categorized into two main categories based on the design methodology employed: Pipeline methods and End-to-end methods~\cite{Chen_2017}. Agents that are developed using conversational AI platforms (task-oriented CAs), such as Rasa, Google Dialogflow, Cognigy, and IBM Watson, fall into the first category. LLM-based CAs such as ChatGPT can be identified as CAs belonging to the second category.
While explicit architectural components can be identified in the pipeline-based CAs, such clear distinctions cannot be identified in end-to-end CAs.

%
%

\subsection{Pipeline-Based Conversational Agents}
\subsubsection{Architecture}
\begin{figure}[!ht]
\centering
  \includegraphics[scale=0.23]{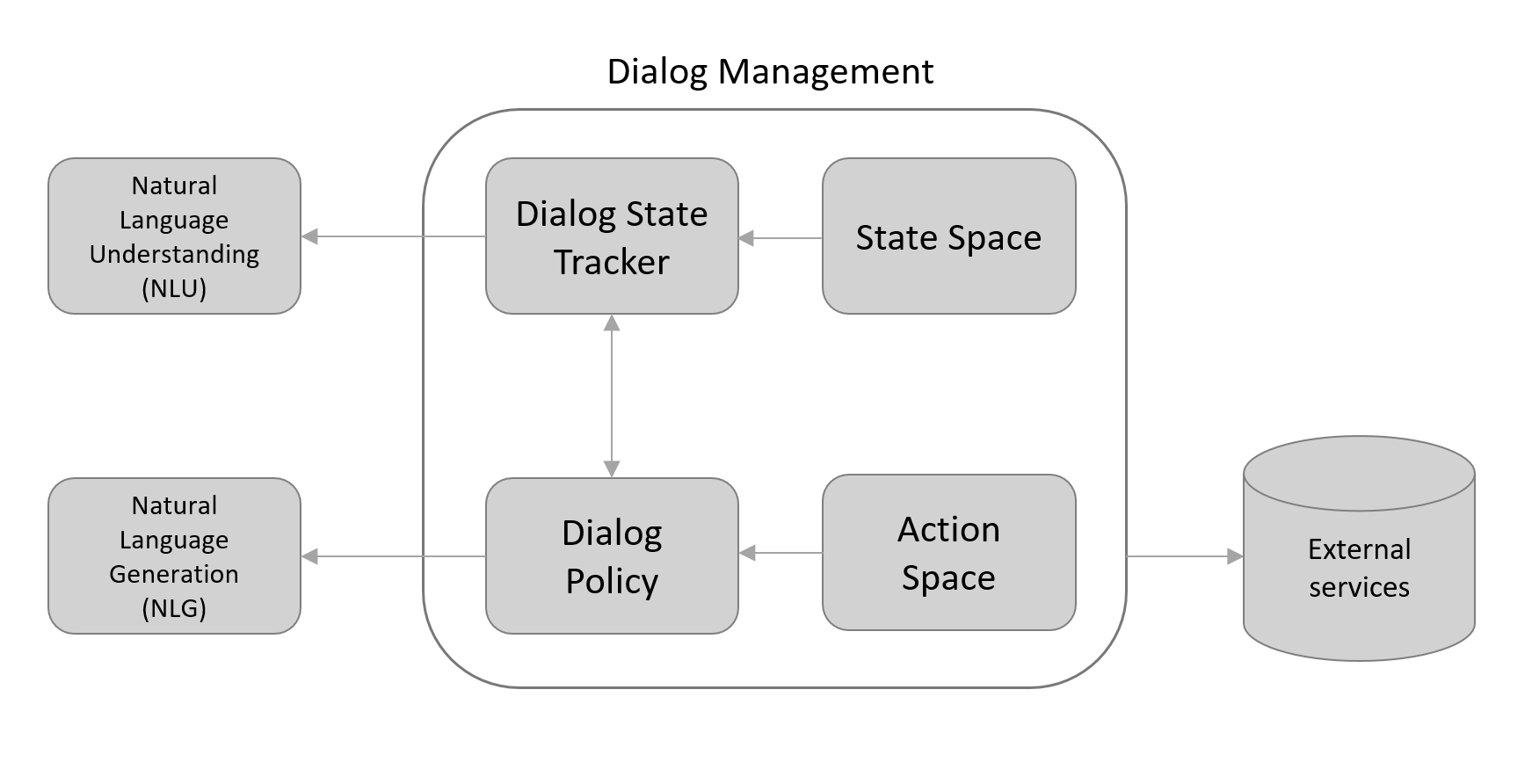}
  \caption{Architecture of a pipeline-based CA. Based on~\cite{Harms.2019,brabra:hal-03626466}}
  \label{fig:architecture}
\end{figure}

In the case of task-oriented CAs, the components that can be explicitly identified are NLU, dialog management, and NLG. A typical architecture of such a CA is shown in Figure~\ref{fig:architecture}.
For NLU and NLG, pipeline-based CAs would traditionally use machine learning-based and template-based approaches, respectively. The dialog management component can be handcrafted, probabilistic, or hybrid. Most of the commercial frameworks and low-code platforms to create task-oriented CAs, such as Google Dialogflow, Cognigy, and IBM Watson, are pipeline-based and use handcrafted rules for dialog management. They are more reliable but less human-like. However, CAs using the probabilistic approach, such as ChatterBot, which are often used for open-domain CAs, create opposite results. Among different platforms, Rasa uses a hybrid approach for the dialog management component~\citep{Harms.2019}.

\subsubsection{Limitations}

Conversational breakdown is a common issue during a conversation with a pipeline-based CA, indicating that the agent did not correctly understand the user’s utterance or responded inadequately to the user's request~\citep{Moore.2019, Flstad.2020}. Conversational breakdowns can lead to frustration, disappointment, and dissatisfaction~\citep{bentley2018understanding,cowan2017can,Luger:2016} if left unaddressed. 
In pipeline-based CAs, these breakdowns occur for various reasons, such as errors during intent and slot recognition, errors during task fulfillment, errors in generating the response, and users' lack of familiarity with a chatbot's intents~\citep{li2020multi}. 
In addition to conversational breakdowns, most of the commercial CAs cannot handle complex queries, lack emotional intelligence, and have limited domain knowledge~\citep{Luo.2022}.

Pipeline-based CAs are also limited regarding the effort in configuring them and how to operate in real-time conversations. In terms of configuration, intent classes, domain entities, and synonym lists need to be created a priori. It requires a certain amount of depth of domain knowledge to come up with suitable notions. The agent's personality and the power dynamics between the agent and user must be defined and expressed by manually creating individual utterances for the bot. The localization to various language varieties requires a significant amount of rework, in particular when it comes to scarcely-supported dialects.

\subsection{Large Language Models}
The advancement of Language Models (LM) in NLP has driven significant progress. In general, the LM aims to predict the next word of a sentence given the current context. With the improvement of research, the concept of the LM has evolved in different stages. From a statistical LM~\citep{jelinek1998statistical}, to predict the next word 
based on Markov assumption, it further progressed through a distributed word representation learning (Word2Vec), which initiated the usage of language model beyond word sequence~\citep{mikolov2013distributed}. Context-aware pre-trained language models~\citep{peters2018deep, devlin2018bert} is one of the early adopters of the modern language model, which sets the paradigm of performing a fine-tuning on any of these pre-trained models on the downstream task and raised the performance achievements on many NLP tasks. One of these models, BERT~\citep{devlin2018bert}, is based on a parallelizable Transformer based network~\citep{vaswani2017attention} with a self-attention mechanism, that begins a new era for future Language models. With the scaling of the model architecture and training data, there has been a rise of many LMs which are named Large Language Model or LLM (e.g., Generative Pretrained Transformer or GPT-series, Pathways Language Model or PaLM-series, etc.)~\citep{brown2020language, chowdhery2022palm, openai2023gpt4, anil2023palm}. One of the prominent differences has been seen in the emergence abilities~\citep{wei2022emergent} where these models could do a series of complex tasks given some specific prompt in a zero-shot or few-shot learning mechanism. Recent models in NLP are based on LLMs and one of the prominent LLM-based conversational agents has been ChatGPT. It is based on the Instruction tuned GPT models (InstructGPT), fine-tuned with Reinforcement learning from human feedback on dialogue data~\citep{ouyang2022training}. An early experiment~\citep{bubeck2023sparks} from the  OpenAI's latest release GPT-4 model, has shown the potential capabilities of the LLMs in different domains that denotes that GPT-4's performance is strikingly close to the human-intelligence level and it is far beyond next-word-level prediction. With the emergence of these LLMs with human-like conversational agents, the evolution of Chatbots has advanced to a different level. 
\subsubsection{LLM-based Conversational Agents}

LLM-based CAs like ChatGPT are trained in a similar way to InstructGPT, specifically optimized for dialogue. Human-generated dialogue data is collected, playing both a Human and an AI role.
They used a three-step process: First, they collected a dataset of human-written demonstrations of the desired model behavior and used it to fine-tune GPT-3 using supervised learning. Next, they collected a dataset of rankings of model outputs and used it to fine-tune the supervised model further using reinforcement learning from human feedback. Finally, they evaluate their models by having human labelers rate the quality of model outputs on a test set. This methodology allows the authors to train language models that are better aligned with user intent and more natural interaction with the inclusion of human-generated data. Recent studies~\citep{longpre2023flan} show the effectiveness of instruct tuning on different LLM models to improve performance on different prompt settings (zero-shot/few-shot settings). One of the recent LLM, Alpaca~\citep{taori2023stanford}, which is built upon self-instruct~\citep{wang2022self} methods on Llama model~\citep{touvron2023llama} shows a 7B parameters LLM can demonstrate high potential to compete with larger GPT-like models. 

\subsubsection{Limitations and Risks}
Despite many benefits of LLMs, they have several limitations~\citep{openai2023gpt4}. For instance, their responses are not reliable (they "hallucinate")~\citep{bang2023multitask, zhao2023survey}. Models like ChatGPT still can produce faithful but nonsensical responses when viewed in the light of the common knowledge in a particular area~\citep{alkaissi2023artificial}. LLMs also have long training times and require huge computation resources; thus, they are not easily obtainable with the latest event knowledge. They do not learn from experience as their context window is limited. For certain task-oriented domains, for example, cybersecurity, the models are unable to assess properly due to context limitations. There are also risks regarding the output of LLM-based models, as they could contain harmful advice, buggy code, or inaccurate information. Like other deep learning models, LLM-based models~\citep{brown2020language} are difficult to interpret due to their complex architecture. Also, their ability to make accurate predictions on new inputs cannot be relied upon, as evidenced by their much higher performance variance than humans on standard benchmarks.




\subsection{Integrating LLM into Pipeline-based Conversational Agents}
As of May 2023, we have found that the involvement of LLMs in pipeline-based CA platforms is mainly limited to NLU and training data generation. For example, Cognigy~\citep{cognigyweb}, with the help of a third-party  Generative AI provider, allows users to generate training data, including intent utterances, lexicons, and flows with pre-configured nodes, Rephrasing bot outputs and completing texts. Even though Cognigy offers a conversation option using generative AI, it is only intended to be used as a preview feature. In another case,~\citep{intentless-rasablog} recently announced the integration of LLMs in their chatbot framework with a new component called IntentlessPolicy. They explain a) how an LLM-based system can take advantage of multiple FAQs without setting up intents for each question, b) how user meaning can be understood in multiple turns of dialogue, and c) how out-of-scope messages can be understandable from the context. They also show that this can be generalized from very little data in a few-shot learning mechanism. They further emphasize that IntentlessPolicy complements intents, rules, stories, and forms. This hybrid approach will better equip with engaging interaction with the user.

%% file: sections/03-approach.tex
\section{LLMs to Overcome Limitations of Pipeline-based CAs}

Despite the various frameworks for building pipeline-based CAs, it still requires substantial time and expertise to design and develop successful CAs. Related tasks concern the design of high-quality training utterances, the definition of intents and consistent and accurate named entities, the selection of domain-specific synonyms, and the localization of training data and responses. Besides, designers must modulate, for instance, training data, dialog management rules, and pre-defined responses to represent desired assistant traits (e.g., client orientation) or personas. We assume that the strengths of LLMs in processing natural language from different countries and domains can substantially shorten at least the time and potentially also the expertise needed to build pipeline-based CAs. Their capability to generate responses matching the style of a generated persona or mimicking an actual person's style could provide new techniques to create attractive CAs.

A second area for improvement is the robustness of a CA at run-time, i.e., when it interacts with a user. Often, pipeline-based CAs produce repetitive responses (robust but less attractive) or experience conversational breakdown because users switched contexts (not robust). In addition, pipeline-based CAs' narrow domain knowledge provokes out-of-scope answers due to smaller training data and limited responses. All of the situations above lower the user's satisfaction and could encourage them to give up on the agent. We assume that LLMs' extensive general and domain knowledge, coupled with their capability of generating attractive and diverse natural language texts, has the potential to achieve more robust and attractive CAs.  

We conclude that LLMs have the capability to enhance pipeline-based CAs during the design and development phase (\emph{delivery accelerator}) and a dialogue with a user (\emph{real-time booster}). In contrast to relying on LLMs only, this hybrid approach is helpful because the pipeline-based approach grants the CA designer more control and transparency over the agent's behavior. The former is critical to counter, for instance, hallucinations, while the latter helps trace and potentially explain unexpected or unwanted behavior.


%% file: sections/04-experiments.tex
\section{GPT-4 Experiments}
To demonstrate the impact of LLMs on CAs, we conducted a series of experiments with GPT-4. The example scenario is a chat agent serving as a client advisor for private banking. A supporting document contains the exact prompts and replies in the conducted experiments.

\paragraph{Parameters} 
We used the Azure Open AI playground with the default parameters for our experiments: Max Response: 800, Temperature: 0.7, Top P: 0.95
Frequency penalty: 0, Presence penalty: 0, Deployment: GPT-4, Past messages included: 10, Max tokens: 8192. The temperature value of 0.7 means that generated responses are not deterministic, i.e., the exact response may vary during reproduction. To keep this article short, we sometimes shorten the actual prompts and answers by inserting an ellipsis.




\subsection{LLM as Delivery Accelerator}
LLM, as a delivery accelerator, involves scenarios to assist developers and designers in building and refining the CA. This can include generating training data, creating lists of entities and synonyms, designing personas to guide the agent's responses, and localizing the agent for different languages and cultures. These tasks can be time-consuming and require significant expertise, so automating them with generative models can save developers time and resources. In Table~\ref{tab:GPT-setup}, we show examples of how LLM can be used in the cases mentioned above for our scenario. The following sections review each development aspect's limitations in pipeline-based CAs and demonstrate how LLMs could address them.

\begin{table*}[ht]
\centering
\begin{tabular}{p{0.3\linewidth} p{0.6\linewidth} }
\hline
\textbf{Benefit} & \textbf{Example} \\
\hline
    Creating intents list & check account balance, transfer funds, pay bill \\
    Generating training utterances for intent classification & Provide a list of different wordings with the intent of canceling an account \\ 
    Creating named entities & account numbers, transaction amounts, date, currency \\
    Creating synonyms lists & insolvent, bankrupt, penniless, broke\\ 
    Traits of good agents & financial knowledge, interpersonal skills, client orientation, ethical standards\\
    Persona-based responses & highly educated client, simple English. \\
    Localization & German, Swiss German, Spanish, French \\
\hline
\end{tabular}
\caption{\label{tab:GPT-setup} Examples of LLMs as a delivery accelerator. A pipeline-based CA for private banking is configured with the help of GPT-4. 
}
\end{table*}

\paragraph{Creating intents lists}
One of the initial steps in designing a pipeline-based CA is to define and identify possible user inquiries or intents. To create a comprehensive list of intents, designers require approaches such as analyzing existing data, sessions with domain experts, and user research. However, LLMs can provide valuable assistance to the designers to gain general insight. We test GPT-4's ability to identify customer intents within a specific industry. We provided the following prompt: \\
{\tt For designing a chatbot, give me a list of 10 most prominent intents in a conversation about banking between a client and an agent.} \\
The first five results provided by GPT-4 (omitting the explanations):\\
{\tt 1. Check account balance ... \\
2. View recent transactions ... \\
3. Transfer funds between accounts ... \\
4. Pay a bill or set up recurring payments ...\\
5. Update personal information ...
}\\
We observe that all of these are common consumer banking interactions that can trigger contact with banking customer service.

\paragraph{Generating training utterances for intent classification}
Writing high-quality training data is a time-consuming task. GPT-4 is capable of helping CA designers with this matter. In~\ref{app:utterances}, we provide ten examples generated by GPT-4. We observe that all generated examples are valid for the intent cancel\_account with various phrasings. However, further prompt tuning would be required to increase variety in length and formality. Generating training data using LLMs and incorporating those into the training pipeline would still require human feedback in the loop to avoid incorrect, biased, or inappropriate training data.

\paragraph{Creating lists of named entities}
We test GPT-4's ability to identify entities relevant to a specific industry. We provided GPT-4 with the following prompt: \\
{\tt For designing a chatbot in the finance domain, give me a list of relevant named entities that the NLP back-end of the chatbot should be able to extract.} \\
GPT-4 returned the following entities (with explanations omitted here): \\
{\tt account numbers, transaction amounts, dates, currency, transaction IDs, percentage rate, financial product names, contact details, company names, bill or invoice numbers, tax-related terms, account types, payment methods, financial goals and financial institutions.} \\
We observe that these are useful terms, although there is an overlap between companies and financial institutions. 

\paragraph{Creating synonyms lists}
We test GPT-4's ability to create synonyms lists for populating entity definitions. We provided the following prompt: \\
{\tt For designing a chatbot in the domain of private banking, give me a synonym list for the word “insolvent”.} \\
GPT-4 returned these 13 terms: \\
{\tt Bankrupt, Impoverished, Penniless, Financially ruined, Broke, Indigent, Destitute, Impecunious, In default, In debt, Insufficient funds, Unable to pay debts, Financially distressed.} \\
While some of these results are unlikely to be used in natural language, all of them are similar in meaning to the term "insolvent". Additionally, the list is not exhaustive, lacking slang terms, e.g., "skint".

\paragraph{Traits of good agents}
We test GPT-4's ability to identify traits associated with a particular role, an important element of chatbot personality design. We provided the following prompt: \\
{\tt Describe the traits of a good client advisor in private banking in max. 100 words.} \\
GPT-4 responded: \\
{\tt A good client advisor in private banking possesses strong financial knowledge, excellent communication and interpersonal skills, and a deep understanding of client's needs and goals. They maintain high ethical standards...} \\
We observe that those are indeed characteristics of a good agent (for details see~\ref{app:goodAgent}).

\paragraph{Persona-based responses}
We show examples of how GPT-4 can be used to personalize responses according to a persona, the lingusitic and societal background of the client. We provided the following prompt: \\
{\tt Write how a chatbot would respond to each of the following scenarios, when talking to a highly educated upper class British client. Use no more than 3 sentences:       
} \\
GPT-4 responded: \\
{\tt 1.	I regret to inform you that the product you've inquired about is no longer available. My sincerest apologies for any inconvenience this may have caused. Should you require any assistance in finding an alternative, please do not hesitate to ask.} \\
We observe that GPT-4 is able to answer this question \emph{in character}, including empathy for the client,  apologies and polite pointing in alternative directions. \\
In a second variation of this experiment, we asked GPT-4 to {\tt write chatbot statements in grammatically correct yet simple English}. GPT-4 responded: \\
{\tt 1.	I'm sorry, but the product you are looking for is no longer available.} \\
The example demonstrates how the language can be adapted by context, which in practice could come from the user data for personalization or from the owner's brand for consistency in customer communication.

\paragraph{Localization}
We test GPT-4's ability to localize agents to different language and dialect varieties of a language (for details see~\ref{app:localization}). We provided the following prompt: \\
{\tt Translate these three statements into German, Swiss German, Spanish, and French. \\
1. I regret to inform you that the product is no longer available. ...
} \\
GPT-4 responded: \\
{\tt 
German: 1.	Es tut mir leid, Ihnen mitteilen zu müssen, dass das Produkt nicht mehr verfügbar ist. ... \\
Swiss German: 4.	Es tuet mer leid, Ihne mitz'teile, dass s'Produkt nümme verfüegbar isch. ... \\
Spanish: 7.	Lamento informarle que el producto ya no está disponible. ... \\
French: 10.	Je regrette de vous informer que le produit n'est plus disponible. ...} \\
We observe that the translations are of high quality in all four languages. Even for Swiss German, a German dialect, the translation is good. This is astonishing because Swiss German does not have a standardized written format for all its varieties. 

\subsection{LLM as Real-time Booster}
LLM, as a real-time booster, involves improving the agent's ability to understand and respond to user input in real-time conversations. This includes contextualizing the conversation to provide more relevant and personalized responses, intent classification and response generation to handle fallback and unhappy paths when the agent is unable to understand or fulfill the user's request, and disambiguating ambiguous requests by asking for more information. Additionally, generative models can assist with integrating closed Q\&A and providing quick and accurate answers to common questions, among others. In Table~\ref{tab:GPT-deploy}, we provide examples of how LLM can improve the above-mentioned aspects in pipeline-based CAs.

\begin{table*}[ht]
\centering
\begin{tabular}{p{0.3\linewidth} p{0.6\linewidth}}
\hline
\textbf{Benefit} & \textbf{Explanation} \\
\hline
    Auto-correct & Correct / rephrase an orthographically and grammatically incorrect utterance to make it more easily classifiable by the bot e.g., "wunt to cancal this accunt" -> "I want to cancel this account" \\
    Context switching & follows the user in switching back and forth between different intents like address change and money transfer \\
    Out-of-scope questions & can be answered when regarding general knowledge \\
    Creating stylistic variability  & utterances can be rephrased, achieving a better writing style while maintaining the same meaning  \\
    Closed Q\&A & Exact formulation of answer is picked from a defined set of options \\
    Summarizing conversation & summarization for hand-over to a human agent  \\
\hline
\end{tabular}
\caption{\label{tab:GPT-deploy}  Examples of LLMs as a real-time booster. A pipeline-based CA can be enhanced during deployment in various ways by GPT-4 overcoming its limitations.} 
\end{table*}

\paragraph{Auto-correction}
We test GPT-4's ability to correct low-literacy utterances. We provided the following examples of phrases which were then auto-corrected by GPT-4. The prompt starts with: \\
{\tt Please rephrase the following 5 utterances into orthographically and grammatically correct American English...} \\
GPT-4 responded: \\
a) {\tt wunt to cancal this accunt $\rightarrow$  \tt I want to cancel this account} \\
b)  {\tt i want 2 get rid of my acount $\rightarrow$  \tt I would like to delete my account}  \\
c) {\tt goodbye to my acount   $\rightarrow$  \tt I am closing my account} \\
We observe that orthographic errors are corrected (a), politeness is introduced (b), and utterances are re-phrased using clear English (c).  

\paragraph{Context Switching} 
We test GPT-4's ability to handle a user that switches between two intents (address change and money transfer) before providing enough information to complete the first task: \\
{\tt User: Hi, I would like to transfer money. My bank account is 334402. \\
Chatbot: Hello! ... Please provide the following information: \\
1.	The recipient's bank account number. \\
2.	The amount you would like to transfer. Once I have this information, I can proceed with the transfer. \\
User: BTW, I also need to change my address: It’s Park Avenue 14. \\
Chatbot: Thank you for providing your new address. To complete the address change, please also provide the following information: \\
1.	City \\
2.	Postal Code ...
\\
User: I live in 10012 New York. \\
Chatbot: Great, thank you for providing the city and postal code for your address change. .... Now, let's get back to the money transfer request. Please provide: 
1.	The recipient's bank account number.
2.	The amount you would like to transfer.... \\
User: I want to transfer 400 Dollars to account number 831123 \\
Chatbot: Thank you for providing the necessary information.} \\
After receiving sufficient information about the address change, the bot gets back to get the complete information for the money transfer.

\paragraph{Out-of-scope Questions}
We added an out-of-scope question to the conversation in the previous paragraph: \\
{\tt User: Where is Germany? \\
Chatbot: Germany is a country located in Central Europe.... } \\
We observe that the bot gives a sensible answer when appending a random general knowledge question to the conversation in the previous paragraph. Staying in context, the agent will then ask the user for a final confirmation: \\
{\tt Chatbot: Before we proceed with the1030
money transfer and address1031
change, please confirm if the1032
provided details are correct...}

\paragraph{Creating stylistic variability}
We test GPT-4's potential to introduce variability in English writing style. We provided the following prompt: \\
{\tt For a chatbot, write 10 variations each one more apologetic than the previous one, of the statement: „I didn't understand what you said, please rephrase." Vary in vocabulary, grammar and tone ...} \\
GPT-4 responded (examples 1, 4, and 9): \\
{\tt 1.	Sorry, I didn't quite get that. Could you rephrase your statement, please? ... \\
4.	Apologies for the confusion, I'm unable to grasp what you're saying. Kindly rephrase your statement.... \\
9.	My most profound apologies for not comprehending your statement. I would be grateful if you could rephrase it for me.} \\
We observe that the generations are of great stylistic variability and that a controlled degree of servitude is introduced into the utterances (see~\ref{app:style} for more details).

\paragraph{Closed Q\&A}
We test  GPT-4's capability to avoid hallucinations in closed Q\&A by only providing exact predefined answers that are not altered (see~\ref{app:closedqa} for details). When testing the system with informally articulated questions, we got five correct answers from 5 trials. 

\paragraph{Summarizing conversation}
We test GPT-4's capability to summarize a conversation between a chatbot and a user and state what the agent picking up the conversation needs to do (see~\ref{app:summ} for details). Summarizing for a CA is particularly useful when requesting confirmation from the client before concluding a conversation or handing over the conversation to a human operator. We observe that the model could deliver the response in the requested format.

%% file: sections/05-outlook.tex
\section{Outlook}
This paper proposes a hybrid approach that leverages LLMs, in particular GPT-4, to enhance pipeline-based CAs. Using this approach, maintainers of existing CAs can adopt new domains and overcome the limitations in conversations with users while ensuring seamless integration with the existing ecosystem. This approach accelerates the CA delivery process through the assistance of LLMs in generating intents, entities, synonyms, respective training data, and agent personality traits.
During deployment, LLMs can boost pipeline-based CAs' performance by utilizing auto-correct, context-switching capabilities, answering out-of-scope questions, creating diverse and stylistically richer responses, and incorporating Closed Q\&A and summarization. This paper presented experiments to showcase the scenarios mentioned above. 

In future work, we will extend the ad-hoc subjective assessment to a more rigorous evaluation among different LLMs and provide an integrated solution to demonstrate the proposed hybrid approach. Given the existing risks regarding the reliability of LLMs~\citep{openai2023gpt4,bang2023multitask, zhao2023survey}, our future research will focus on examining the factors that prompt business owners to consider the integration of LLMs within their pipeline-based CAs.

%% file: sections/06-appendix.tex
\section{Appendix}
\label{sec:appendix}

This appendix contains selected transcripts of the prompts and the generated responses from GPT4.

\subsection{Generating training utterances for intent classification}
\label{app:utterances}
Prompt: \\
{\tt Write 10 varied utterances to train a chatbot intent called cancel\_account, which is for customers who want to close their account with  ABC Bank. Sometimes mention the bank name, sometimes don’t.} \\
Output: \\
{\tt 1.	I would like to close my account with ABC Bank, please help me with the process. \\
2.	Can you please guide me on how to cancel my account at ABC Bank? \\
3.	I want to terminate my banking relationship with ABC Bank, how can I do that? \\
4.	I'm thinking of closing my account, what is the procedure? \\
5.	I've decided to cancel my ABC Bank account, can you assist me with this? \\
6.	Please help me shut down my account with your bank. \\
7.	I no longer need my account at ABC Bank, how can I close it? \\
8.	What's the process to deactivate my account with ABC Bank? \\
9.	I would like to cancel my account; can you guide me through the steps? \\
10.	I need to close my bank account, what information do you need from me?}

\subsection{Traits of a good agent}
\label{app:goodAgent}
Prompt: \\ 
{\tt Describe the traits of a good client advisor in private banking in max. 100 words.} \\
Output: \\
{\tt A good client advisor in private banking possesses strong financial knowledge, excellent communication and interpersonal skills, and a deep understanding of clients' needs and goals. They maintain high ethical standards, practice discretion and confidentiality, and build long-lasting relationships based on trust. Additionally, they demonstrate adaptability, staying updated on market trends and regulations, and proactively identify opportunities to grow clients' wealth. They are also highly organized, adept at problem-solving, and excel at collaborating with various stakeholders to deliver tailored financial solutions.}

\subsection{Localization}
\label{app:localization}
Prompt: \\
{\tt Translate these three statements into German, Swiss German, Spanish and French. \\
1. I regret to inform you that the product is no longer available. \\
2. I implore you to reconsider cancelling your account. \\
3. I shall now direct you to an agent who can further assist you.} \\
Output: \\
{\tt German: 1.	Es tut mir leid, Ihnen mitteilen zu müssen, dass das Produkt nicht mehr verfügbar ist. \\
2.	Ich bitte Sie inständig, die Kündigung Ihres Kontos zu überdenken. \\
3.	Ich werde Sie jetzt an einen Agenten weiterleiten, der Ihnen weiterhelfen kann. \\
Swiss German: 4.	Es tuet mer leid, Ihne mitz'teile, dass s'Produkt nümme verfüegbar isch. \\
5.	Ich bitte Sie inständig, d'Chündigung vo Ihrem Konto z'überdenke. \\
6.	Ich wird Sie jetzt zuenem Agent wyterleite, wo Ihnen cha wyterhelfe. \\
Spanish: 7.	Lamento informarle que el producto ya no está disponible. \\
8.	Le imploro que reconsidere cancelar su cuenta. \\
9.	Ahora le dirigiré a un agente que pueda asistirle más. \\
French: 10.	Je regrette de vous informer que le produit n'est plus disponible. \\
11.	Je vous implore de reconsidérer l'annulation de votre compte. \\
12.	Je vais maintenant vous diriger vers un agent qui pourra vous aider davantage.}

\subsection{
Creating stylistic response variability}
\label{app:style}
Prompt: \\
{\tt For a chatbot, write 10 variations each one more apologetic than the previous one, of the statement: „I didn't understand what you said, please rephrase.“ Vary in vocabulary, grammar and tone, including terse, sincere and apologetic variations.} \\
Output: \\
{\tt 1.	 Sorry, I didn't quite get that. Could you rephrase your statement, please? \\
2.	My apologies, I'm having trouble understanding. Would you mind rephrasing your question? \\
3.	I'm sorry, I didn't comprehend your message. Please rephrase it for me. \\
4.	Apologies for the confusion, I'm unable to grasp what you're saying. Kindly rephrase your statement. \\
5.	I deeply regret that I didn't understand your message. Please accept my apologies and rephrase your question. \\
6.	My sincerest apologies, I'm struggling to comprehend your message. Could you kindly restate it for me? \\
7.	I'm terribly sorry for not understanding your words. Please forgive me and rephrase your statement. \\
8.	I feel so apologetic for being unable to understand what you said. Please give me another chance and rephrase your message. \\
9.	My most profound apologies for not comprehending your statement. I would be grateful if you could rephrase it for me. \\
10.	I am extremely sorry for my inability to understand your message. It would mean a lot if you could kindly rephrase it for me.}

\subsection{Closed Q\&A}\label{app:closedqa} 
Prompt: \\
{\tt For each question literally answer one of the below answers in exactly that wording, if those answers are suitable. If none of the below answers are a suitable answer to the question answer: “Please call 001 23 45 89 01”. \\
1.	To change your address you need to sent a mail to info.company.com including your new and old complete address. \\
2.	If you want to open a bank account, provide a copy of your password and a list of current bank accounts. \\
3.	If you want to close an account call 001 23 45 89 28 \\
4.	To change your password sent a mail to info.company.com with that request.} \\
Output: \\
{\tt User: I have a new address \\
Chatbot: To change your address you need to sent a mail to info.company.com including your new and old complete address. \\
User: How can I get an account with your company? \\
Chatbot: If you want to open a bank account, 
provide a copy of your password and a list of current bank accounts. \\
User: I want to quit \\
Chatbot: If you want to close an account call 001 23 45 89 28 \\
User: I forgot my pwd \\
Chatbot: To change your password sent a mail to info.company.com with that request. \\
User: What are the interest rates I need to pay for a mortgage? \\
Chatbot: Please call 001 23 45 89 01.
}

\subsection{Summarizing conversation}
\label{app:summ}
Prompt: \\
{\tt Summarise the following conversation between a chatbot and a person, and state what the agent picking up the conversation needs to do. \\
--- \\
Chatbot: Hi, how can I help? \\
User: I need to get a new debit card \\
Chatbot: I can help you order a new debit card. Is this a new card or a replacement? \\
User: Replacement \\
Chatbot: Is your current card lost, damaged or stolen? \\
User: Damaged \\
Chatbot: Please go to www.cardreplace.com to request your new card. Did I help you today? \\
User: The trouble is the address you have for me is out of date, so before you post it you need to update my address \\
Chatbot: I’m worry I didn’t understand that. Did I help you today? \\
User: I need to update my address \\
Chatbot: My colleague can help you this query, I’m connecting you now. Feel free to add any information that might be help now. \\
User: My new address is 1 Main Street, Capital City, Countryland, AA1 XZY. \\
---\\
Use this format: \\
Agent Action Required: \\
Summary: } \\
Output: \\
{\tt Agent Action Required: Update the user's address and assist with the debit card replacement process. \\
Summary: The user needs a replacement debit card due to a damaged one. However, their address on file is outdated. The user provided their new address as 1 Main Street, Capital City, Countryland, AA1 XZY.}